\documentclass[conference]{IEEEtran}
\ifCLASSINFOpdf
   \usepackage[pdftex]{graphicx}
   \DeclareGraphicsExtensions{.pdf,.jpeg,.png}
\else
   \usepackage[dvips]{graphicx}
   \DeclareGraphicsExtensions{.eps}
\fi
\usepackage[cmex10]{amsmath}
\begin{document}
\title{An Efficient Architecture for Predicting the Case of Characters using Sequence Models}

\author{\IEEEauthorblockN{Gopi Ramena, Divija Nagaraju, Sukumar Moharana, Debi Prasanna Mohanty, Naresh Purre}
\IEEEauthorblockA{Samsung R\&D Institute\\
Bangalore, India\\
\{gopi.ramena, d.nagaraju, msukumar, debi.m, naresh.purre\}@samsung.com}
}


%Telephone: (800) 555--1212, Fax: (888) 555--1212}


\maketitle

\begin{abstract}
%\boldmath
The dearth of clean textual data often acts as a bottleneck in several natural language processing applications. The data available often lacks proper case (uppercase or lowercase) information. This often comes up when text is obtained from social media, messaging applications and other online platforms. This paper attempts to solve this problem by restoring the correct case of characters, commonly known as Truecasing. Doing so improves the accuracy of several processing tasks further down in the NLP pipeline. Our proposed architecture uses a combination of convolutional neural networks (CNN), bi-directional long short-term memory networks (LSTM) and conditional random fields (CRF), which work at a character level without any explicit feature engineering. In this study we compare our approach to previous statistical and deep learning based approaches. Our method shows an increment of 0.83 in F1 score over the current state of the art. Since truecasing acts as a preprocessing step in several applications, every increment in the F1 score leads to a significant improvement in the language processing tasks.
\end{abstract}

%\keywords{neural networks, sequence models, deep learning, capitalization, truecase}

% \fi
\IEEEpeerreviewmaketitle

\section{Introduction}
The advent of the internet in the 1900s led to a surge in the amount of accessible data. The abundance of websites, blogs and newsletters led to more and more textual data being processed everyday. This fuelled the emergence of the relatively new field of natural language processing. In more recent times, the number of people contributing to this pool of data has increased. The use of social media platforms and messaging applications has modified our perception of language. There is a prevalent disregard for punctuation and casing in an attempt to put forth our views across faster. A result of this is that a huge percentage of data is deemed unusable for natural language processing. By neglecting such a significant portion of information, we are hindering the progress of NLP applications which must evolve constantly according to our changing language. 

In this paper, we sought to tackle a significant subproblem of the task of converting grammatically and semantically incorrect data into a usable format, namely, casing. Truecasing \cite{lita2003truecasing} refers to converting letters into their correct case (upper or lower) taking semantic and syntactic criteria into consideration. The task often removes ambiguity in sentences. For example, consider the word "bill" in the sentences "Jim forgets to pay his bill on time" and "Jim invited Bill to his party". The former usage refers to a written statement of money, whereas the latter is an individual's name.  Such discrepancies make it difficult to perform automated text processing because disambiguation ability is absent in machines. 

Truecasing is important when dealing with text obtained from chats, tweets and other prevalent forms of informal communication. Users seldom use the correct case on these platforms. While working on text processing applications customised for these such as topic modelling on chat data or performing named entity recognition on twitter data incorrect casing is often a hindrance. 

In this paper we propose a model which uses CNNs \cite{lecun1989backpropagation} for generating character-level representations and a bi-directional LSTM for modelling sequence information. Furthermore, the final output tagging is done with the help of CRF decoding. We evaluate and analyze our methodology and compare the performance to various other methods. In Section~\ref{sec:related} we discuss several approaches to truecasing taken before.  Section~\ref{sec:arch} and ~\ref{sec:res} describe our network architecture and experimental results respectively. We finally conclude with a discussion of possible applications and future work.

\section{Related Work} \label{sec:related}
Truecasing is a relatively unexplored subject and previous studies on it have vastly different perspectives on this subtask. Several previous works consider truecasing as a subset of word sense disambiguation (WSD). WSD seeks to identify in what sense a particular word is used in a sentence. Hence variations in the casing of different words can just be considered as different senses in which the word can be used. WSD is often tackled by taking contextual information into account. The same approach can be used for truecasing too. Many studies also view truecasing as a special case of spelling correction. There has been vast literature on spell-checking \cite{spellingcorrection2}. When modifying spell-checking for truecasing one needs to merely consider the uppercase and lowercase characters as separate entities and train the model with this modification. This approach may not be completely reliable due to lexical and morphological considerations. Few studies focused on Named Entity Recognition (NER) \cite{casingforner} and proper noun identification \cite{idpropername} specifically include casing as part of the larger problem and tackle it in a more customized way. Efficient casing information is an integral part of proper noun identification. Brill and Moore \cite{noisy} also apply a noisy channel model to spelling correction and look at uncased characters as noise which must be eliminated.

Truecasing has been attempted using different tools for many decades, even though the term truecasing was coined much later \cite{lita2003truecasing}. Studies in the late 1900s on casing information used rule based criteria to determine the appropriate case for a character. This included simple rules like checking for full stops and adding capital letters after it and using a word bank of proper nouns. Lita et al. \cite{lita2003truecasing} revived the study of case information and named the process "tRuEcasIng". The paper proposes a statistical and language modelling based truecaser. More importantly, the study emphasizes the need for truecasing as a building block in NLP applications. Despite being the benchmark approach for several years, statistical methods have a disadvantage of not performing well on unseen data. A recent study \cite{susanto2016learning} uses recurrent neural networks for truecasing and compares it with statistical approaches.
The deep learning approaches on truecasing can be divided into character and word level. Word based truecasing converts each word into its most frequently used form. Approaches include Hidden Markov Model (HMM) based taggers \cite{hmmforcasing}, discriminative taggers and CRFs. Statistical methods also inevitably work at a word level. Raymond et al. \cite{susanto2016learning} is the first study to work at a character-level and provides detailed comparison between character and word level methods.

\section{Neural Network Architecture} \label{sec:arch}
In this section we provide a description of our proposed architecture to truecasing using character-level RNN. Our approach uses CNN to generate character representations from the input sentence which are then passed to a 2 layered Bi-LSTM. The bidirectional LSTM helps in getting the contextual information from adjacent characters. This output is further decoded by a CRF layer to get the final sequence of the proper case information(upper or lower).

\subsection{CNN for Character-Level Embeddings}
Convolutional Neural Networks (CNNs) \cite{lecun1989backpropagation} have been shown to be an effective approach to extract morphological information from characters in a sequence and encode it into neural representations. We use only character embeddings as the inputs to the CNN, without character type features. We use a 1D convolution layer with kernel size of 5 and 32 filters. We use "same" padding and a ReLU activation in the CNN layer.

A dropout layer \cite{srivastava2014dropout} is applied to the inputs before passing it to the CNN. An input dropout of 0.25 was found to perform best in our experiments.

\subsection{Bi-directional LSTM}

\subsubsection{Long Short-term Memory Networks(LSTMs) Unit}
Recurrent neural networks (RNNs) are a family of neural networks that capture time dynamics on sequential data. They take a sequence of vectors as input and return an output sequence which is a representation of information at every input step. In theory, RNNs are capable of capturing long-distance dependencies, but in practice they fail to do so due to the problem of vanishing or exploding gradients \cite{bengio1994learning, pascanu2012understanding}.

Long Short-term Memory Networks (LSTMs) \cite{hochreiter1997long} are variants of RNNs designed to tackle this issue by incorporating a memory cell and have been shown to capture long-range dependencies. An LSTM unit is composed of three multiplicative gates which control the proportions of information to forget and to pass on to the next time step. The content of the memory cell is updated additively and multiplicatively, mitigating the vanishing gradients problem in vanilla RNNs.

\subsubsection{Bi-LSTM}
An LSTM computes representation of the left (past) context of the sequence at every input t. For many sequence labeling tasks, another backward LSTM that reads the same sequence in reverse, is found to add useful information by generating the right (future) context. These are two distinct networks with different parameters. This forward and backward LSTM pair is referred to as a bidirectional LSTM \cite{graves2005framewise}.
The idea is to capture both past and future information in the two separate hidden states. The forward LSTM computes a representation $\displaystyle \overrightarrow{h_{t}}$ of the left context of the sentence at every character t. Similarly, the backward LSTM computes a representation of the right context $\displaystyle \overleftarrow{h_{t}}$. Then both the left and right context representations are concatenated to form the final character representation, $\displaystyle h_{t} =\left[\overrightarrow{h_{t}} \ ;\ \overleftarrow{h_{t}}\right]$.

For our experiment we use a 2 layer Bi-LSTM model with 150 hidden nodes. We also add a recurrent dropout of 0.25 in the Bi-LSTM layer \cite{srivastava2014dropout}.

\subsection{CRF Decoding}
Conditional Random Fields (CRF) \cite{lafferty2001conditional} is a sequence modelling method often used for structured prediction of sequences. For labeling of sequences, it is often helpful to consider relations between labels in neighboring sequences and make joint label tagging decisions for the input sequence. Without the CRF layer, the model makes independent classification decisions from the output layer to determine the case of the predicted character. Therefore instead of predicting tag labels independently, we model them jointly as a sequence prediction using a conditional random field \cite{sang1999representing}.

We define an input sequence $\displaystyle X$, and the sequence of labels corresponding to $\displaystyle X$ is defined by $\displaystyle Y$.

The probability defined by the sequence CRF model for all possible label sequences $\displaystyle Y$ given $\displaystyle X$ is defined by the following conditional probability:
\begin{equation}
p( Y|X) \ =\ \frac{e^{s( X,Y)}}{\sum\limits _{y'\in \tilde{Y}} e^{s( X,y')}}
\end{equation}
where $\displaystyle s(X,Y)$ is the potential score function defined by
\begin{equation*}
s( X,Y) \ =\ \sum ^{n}_{i=0} A_{y_{i} ,\ y_{i+1}} +\sum ^{n}_{i=1} P_{i,\ y_{i}}
\end{equation*}
where $\displaystyle A_{i,j}$ denotes the transition scores of a transition from tag $\displaystyle i$ to tag $\displaystyle j$, and $\displaystyle P$ denotes the matrix of output scores by the Bi-LSTM network, and $\displaystyle \tilde{Y}$ denotes all possible label sequences for input sequence $\displaystyle X$.

For training of the CRF, we use the maximum conditional log likelihood estimation. During decoding, we predict the sequence of outputs with the highest conditional probability, which is given by:
\begin{equation}
y^{*} \ =\ \underset{y\in \tilde{Y}}{argmax} \ p( Y|X)
\end{equation}

\subsection{CNN-Bi LSTM-CRF Neural Network}
\begin{figure}
  \centering
  \includegraphics[width=\linewidth]{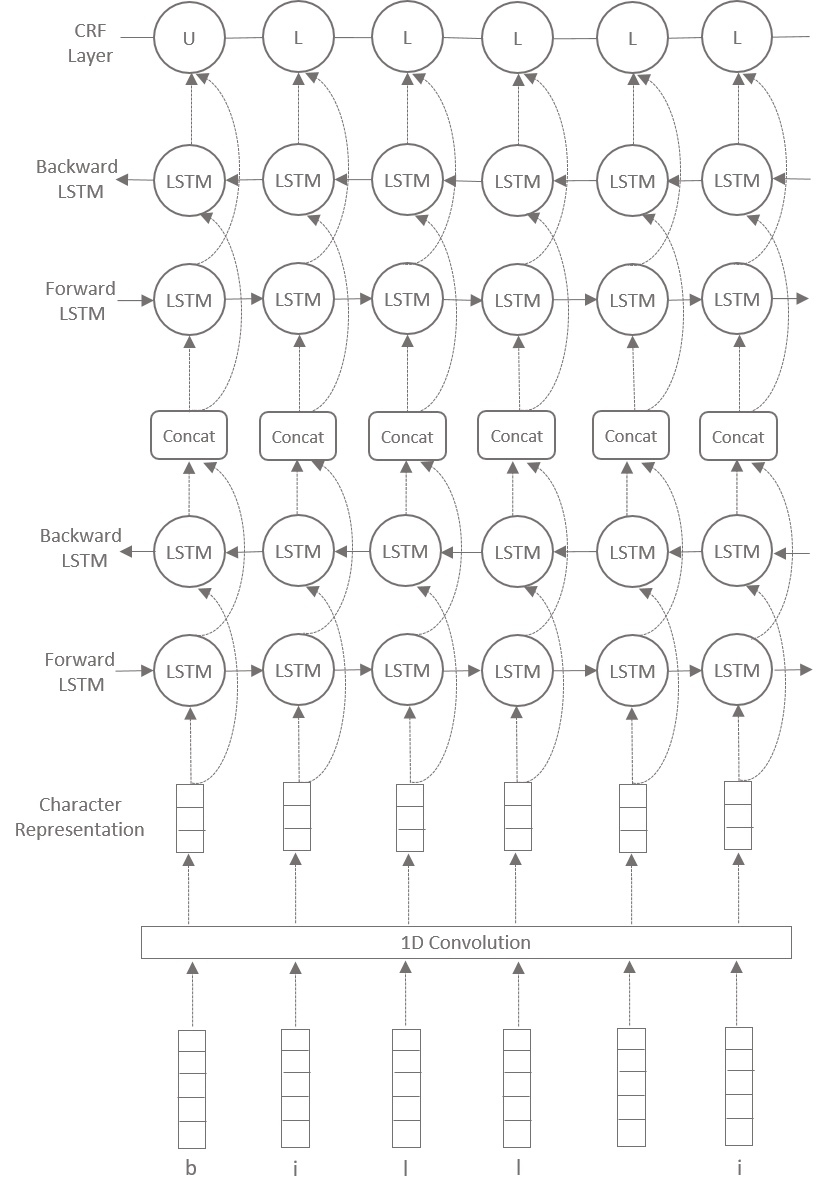}
  \caption{The main architecture of our neural network}
  \label{fig:network}
\end{figure}
Finally, we construct our neural network model by combining the components described previously. Figure~\ref{fig:network} illustrates the architecture of our neural network in detail. Our model architecture consists of a CNN layer to get the character representations. For each character, the character embedding vector is input to the 1 dimensional CNN to get the character representations. The use of CNN has been shown to be efficient in extracting contextual character representations from text \cite{santos2014learning, chiu2016named}. Then this character-level representations are fed to the 2 layered Bidirectional LSTM. We use a Bi-LSTM instead of an LSTM as it extracts contextual information from both directions of the sequence. For example, the probability of the next character being upper case is more after a full stop. The character-level \cite{ballesteros2015improved} Bi-LSTM helps in capturing these contexts which help improve the results significantly. The output vectors of the Bi-LSTM are concatenated and a softmax classifies it to one of upper (U) or lower (L) case classes. The CRF layer jointly decodes the best final sequence of character case in the input sentence to get the final output of the correct case. The CRF layer has been successful in various sequence problems like POS tagging \cite{ma2016end} and Named Entity Recognition \cite{ma2016end, lample2016neural}.

These final output vectors are fed into the CRF layer to jointly decode the best final sequence of character case in the input sentence.

\section{Experiments and Results} \label{sec:res}

\subsection{Dataset}
We evaluate our neural network model on the Wikipedia text simplification dataset from \cite{coster2011simple}. We use the same train/dev/test splits as in the original paper and provided dataset. The input test data is lowercased. Most previous works did not evaluate their approaches on the same dataset. Raymond et al. \cite{susanto2016learning} used this same dataset to produce state of art results. We evaluate Lita et al. \cite{lita2003truecasing} on this same dataset and list down F1 scores of other previous works on this dataset. The corpora statistics are shown in Table ~\ref{tab:corpus}. We did not perform any preprocessing for the sentences in the corpora, leaving our system truly end-to-end.

\begin{table}
  \caption{Wiki Dataset distribution}
  \label{tab:corpus}
  \centering
  \begin{tabular}{lllll}
  \hline
    \bfseries Split & \bfseries No. of words & \bfseries No. of chars\\
  \hline
    train & 2.9M & 16.1M\\
    dev & 294K & 1.6M\\
    test & 32K & 176K\\
  \hline
  \end{tabular}
\end{table}

\subsection{Network Training}
For our approach presented, we train the model using mini-batch stochastic gradient descent \cite{li2014efficient} with batch size of 64 and Adam Optimizer \cite{kingma2014adam} with a learning rate of 0.002. We use dropout regularization \cite{srivastava2014dropout} with 0.25 input dropout probability. The CNN layer consists of a kernel size of 5 and 32 filters, thus creating our character representation of 32 dimensions. Our Bi-LSTM consists of 2 layers and 150 hidden nodes with a recurrent dropout of 0.25 probability.

We use a softmax at the end with 2 output labels, upper (U) or lower (L), which is decoded at the final CRF layer. We experimented with different output layers like using softmax with number of labels as vocab size and taking the maximum probability character among the upper or lower cases of the next character. We also tried predicting the next character from the learnt model and taking its case as the case of the next known character. The best performance was observed with our current output layer having a softmax classifier with upper or lower labels. Adding the CRF layer further increased the model F1 score.

\subsection{Results}

\begin{table}
  \caption{Truecasing performance comparison}
  \label{tab:comparison}
  \begin{tabular}{lllll}
  \hline
    & \bfseries Accuracy & \bfseries Precision & \bfseries Recall & \bfseries F1\\
  \hline
    Lita et al. \cite{lita2003truecasing} & 94.93 & 89.42 & 84.41 & 86.84\\
    Stanford CoreNLP \cite{manning2014stanford} & 96.60 & 94.96 & 87.16 & 90.89\\
    Raymond et al. \cite{susanto2016learning} & 97.41 & 93.72 & 92.67 & 93.19\\
    CNN + Bi-LSTM + CRF & 97.84 & 94.92 & 93.14 & \bf{94.02}\\
  \hline
  \end{tabular}
\end{table}

Table ~\ref{tab:comparison} presents our results in comparison to the previous works by Lita et al. \cite{lita2003truecasing}, Raymond et al. \cite{susanto2016learning} and also the TrueCaseAnnotator in Stanford CoreNLP \cite{manning2014stanford}. We outperform all previous methods and obtaining an improvement of 0.83 F1 score from Raymond et al. \cite{susanto2016learning}. Additionally, our model does not use any hand-engineered features or word case lookup dictionaries. We also present detailed results of our various experiments in table ~\ref{tab:model_scores}. 

\begin{table}
  \caption{Performance of our model}
  \label{tab:model_scores}
  \begin{tabular}{lllll}
  \hline
    & \bfseries Accuracy & \bfseries Precision & \bfseries Recall & \bfseries F1\\
  \hline
    Bi-LSTM & 97.40 & 94.55 &91.73 & 93.12\\
    CNN + Bi-LSTM & 97.58 & 93.72 & 93.31 & 93.53\\
    CNN + Bi-LSTM + CRF & 97.84 & 94.92 & 93.14 & \bf{94.02}\\
  \hline
  \end{tabular}
\end{table}

\section{Conclusion and Future Work}
In this paper, we have discussed a novel model architecture for truecasing which outperforms the current state of the art. Our model uses a combination of CNN and Bi-LSTM layers with a final CRF layer to get the desired output. Additionally, we have compared our methodology to previous approaches to the problem, which include statistical and deep learning methods.

There can be several potential directions for future work. Firstly, truecasing aids in various NLP tasks, such as Named Entity Recognition and Machine Translation. Proper capitalization allows easier detection of proper nouns, which are the starting points of these NLP tasks. Some translation systems use statistical machine learning techniques, which could make use of the correct case information to increase accuracy.  Since our model does not require any explicit feature engineering, its addition to different architectures should be seamless. Truecasing can be particularly useful in Optical Character Recognition (OCR) systems. OCR systems often suffer from a diminished accuracy due to mistakenly identifying similar letters, such as $\textit{"p"}$ as a $\textit{"P"}$ and vice versa. When contextual information is incorporated, casing can be identified and discrepancies prevented. Another interesting direction to pursue would be automated cleaning of the text obtained from tweets and short messages. This would greatly improve the performance of other tasks further down in the NLP pipeline. The impact of the addition of truecasing as a separate module in current text processing applications remains to be seen.


%\section*{Acknowledgment}



\bibliographystyle{IEEEtran}
\bibliography{IEEEabrv,references}
%\begin{thebibliography}{1}


%\end{thebibliography}

\end{document}